\documentclass[conference,letterpaper]{IEEEtran}
\renewcommand{\normalsize}{\fontsize{11}{13}\selectfont}
\IEEEoverridecommandlockouts
\usepackage[top=0.8in, bottom=1.05in, left=0.65in, right=0.65in]{geometry}

\usepackage{cite}
\usepackage{amsmath,amssymb,amsfonts}
\usepackage{algorithmic}
\usepackage{graphicx}
\usepackage{textcomp}
\usepackage{xcolor}
\def\BibTeX{{\rm B\kern-.05em{\sc i\kern-.025em b}\kern-.08em
    T\kern-.1667em\lower.7ex\hbox{E}\kern-.125emX}}

\usepackage{algorithm}
\usepackage{multirow}
\usepackage{bbding}
\usepackage{array}
\usepackage{makecell}
\usepackage{amsmath}
\usepackage{xspace}
\usepackage{color}
\usepackage{amssymb}
\usepackage{booktabs}
\usepackage{pifont}
\usepackage{bm}

\usepackage{newfloat}
\usepackage{listings}
\def\eg{\textit{e.g.}\xspace}
\def\ie{\textit{i.e.}}
\def\etal{\textit{et al.}\xspace}

\def\name{\xspace\textbf{MICACL}\xspace}

\begin{document}

\title{MICACL: Multi-Instance Category-Aware Contrastive Learning for Long-Tailed Dynamic Facial Expression Recognition
\thanks{† Corresponding Author.}
}

\author{
\IEEEauthorblockN{Feng-Qi Cui\textsuperscript{1,3}, Zhen Lin\textsuperscript{2}, Xinlong Rao\textsuperscript{1}, Anyang Tong\textsuperscript{2}, Shiyao Li\textsuperscript{2}, Fei Wang\textsuperscript{2,3}, Changlin Chen\textsuperscript{1}, Bin Liu\textsuperscript{1,†} }

\IEEEauthorblockA{\textit{\textsuperscript{1}University of Science and Technology of China,}
Hefei, China \\
\textit{\textsuperscript{2}Hefei University of Technology,}
Hefei, China \\
\textit{\textsuperscript{3}IAI, Hefei Comprehensive National Science Center,}
Hefei, China \\
Email:\textsuperscript{†}flowice@ustc.edu.cn }

}

\maketitle

\begin{abstract}
Dynamic facial expression recognition (DFER) faces significant challenges due to long-tailed category distributions and complexity of spatio-temporal feature modeling. While existing deep learning-based methods have improved DFER  performance, they often fail to address these issues, resulting in severe model induction bias. To overcome these limitations, we propose a novel multi-instance learning framework called \name,  which integrates spatio-temporal dependency modeling and long-tailed contrastive learning optimization. Specifically, we design the Graph-Enhanced Instance Interaction Module (GEIIM) to capture intricate spatio-temporal between adjacent instances relationships through adaptive adjacency matrices and multiscale convolutions. To enhance instance-level feature  aggregation, we develop the Weighted Instance Aggregation Network (WIAN),  which dynamically assigns weights based on instance importance. Furthermore, we introduce a Multiscale Category-aware Contrastive Learning (MCCL) strategy to balance training between major and minor categories. Extensive experiments on in-the-wild datasets (\ie, DFEW and FERV39k) demonstrate that \name achieves state-of-the-art performance with superior robustness and generalization.
\end{abstract}


\section{Introduction}
\label{sec:intro}

Dynamic Facial Expression Recognition (DFER) has made remarkable progress in recent years, transitioning from relying on handcrafted features, such as FACS+ \cite{FACS} and HOG-3D \cite{HOG3D}, to deep learning-based methods\cite{10339891,10587029,10899398,10916982}. Although traditional methods perform well in controlled environments, they heavily rely on expert knowledge and struggle to adapt to dynamic and complex scenarios. With the introduction of datasets such as DFEW \cite{dfew}, DFER research has increasingly shifted toward "in-the-wild" conditions, where facial expression data are more natural and diverse, reflecting the demands of real-world applications such as intelligent human-computer interaction \cite{10804189,10699348,10090421,10508334} and mental health monitoring \cite{wang2024facialpulse,cui2025learningheterogeneitygeneralizingdynamic,feng2025rf,10149418}.

\begin{figure}[t]
\centering
\includegraphics[width=0.46\textwidth]{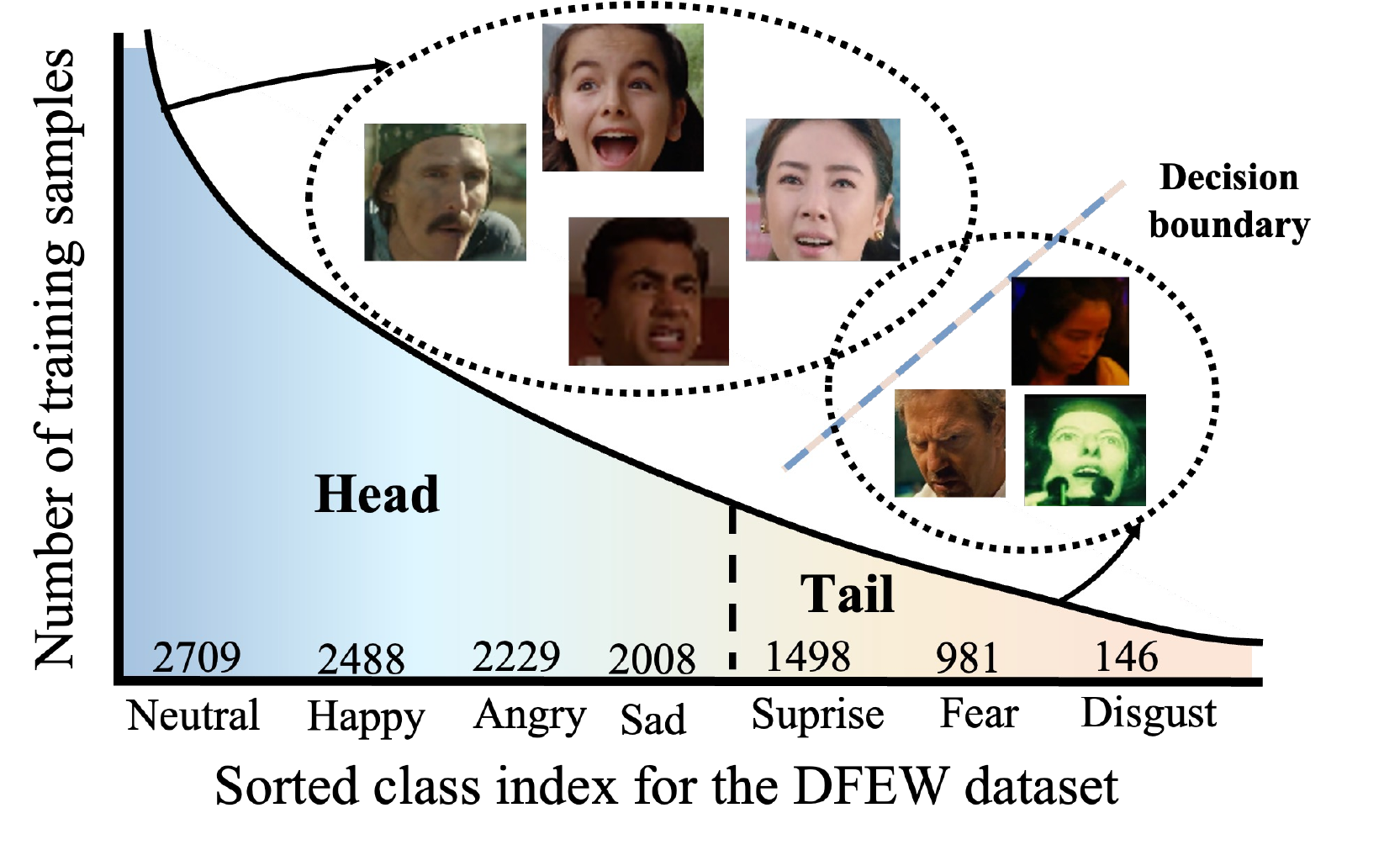} 
\caption{The long-tail data label distribution in DFEW dataset, the feature space learned on these samples is usually larger for the head category than for the tail category, and the decision boundary is usually biased towards the dominant category.}
\label{fig1}
\end{figure}

However, "in-the-wild" conditions bring significant challenges,  particularly the widespread presence of long-tailed distributions in dynamic facial expression data. As shown in Fig. \ref{fig1}, tail emotional categories contain far fewer samples compared to head categories. This imbalance not only constrains the model's capacity to recognize expressions from tail categories,  but also exacerbates overfitting to head-category features and reduces intercategory feature discrimination \cite{10105457,9613773,9075376}. 
Additionally, the high diversity and unconstrained nature of "in-the-wild" data make it even more challenging to capture spatio-temporal dependencies in dynamic sequences, making it difficult to accurately identify critical frames and inter-frame relationships \cite{10204167}. These factors significantly increase the difficulty of achieving robust DFER.
To address these challenges, we propose a novel framework named\name, which deeply integrates spatio-temporal relationship interaction modeling and long-tailed multiscale contrastive optimization to achieve robust and balanced DFER performance improvements.
First, we design the Graph-Enhanced Instance Interaction Module (GEIIM), which constructs complex inter-frame relationships through adaptively generated adjacency matrices and employs multiscale convolutions to enable spatio-temporal feature interactions. 
For instance feature aggregation, we develop the Weighted Instance Aggregation Network (WIAN), which dynamically assigns weights based on the importance of different instances in the input sequence.   This enhances the representation of critical frames while suppressing irrelevant or redundant information during global feature aggregation.   Additionally, in the optimization phase, we introduce a novel Category-Aware Multiscale Contrastive Learning (CAMCL) strategy that dynamically balances training weights between head and tail categories, focusing specifically on tail-category features.   This approach allows the model to adapt to long-tailed distributions while maintaining accurate recognition of major categories.   In general, our main contributions are summarized as follows:

\begin{itemize}
\item To the best of our knowledge, this is the first work to leverage the inherent category distribution in dynamic facial expressions for feature contrast across multiscale feature spaces, significantly enhancing the network's ability to learn long-tail categorical features in expressions.
\item We introduce \name, a new multi-instance learning framework featuring adaptive instance interaction (GEIIM) and dynamic long-instance aggregation (WIAN). This framework is designed to achieve high-speed, high-accuracy, and high-robustness DFER.
\item We conduct extensive experiments on two prominent in-the-wild DFER datasets (i.e., DFEW and FERV39k) to demonstrate the superiority of our approach in long-tailed dynamic facial expression learning and multi-instance modeling. \name achieves excellent performance in terms of recognition accuracy, computational efficiency, and category boundary delineation.

\end{itemize}

\begin{figure*}[t]
\centering
\includegraphics[width=0.89\textwidth]{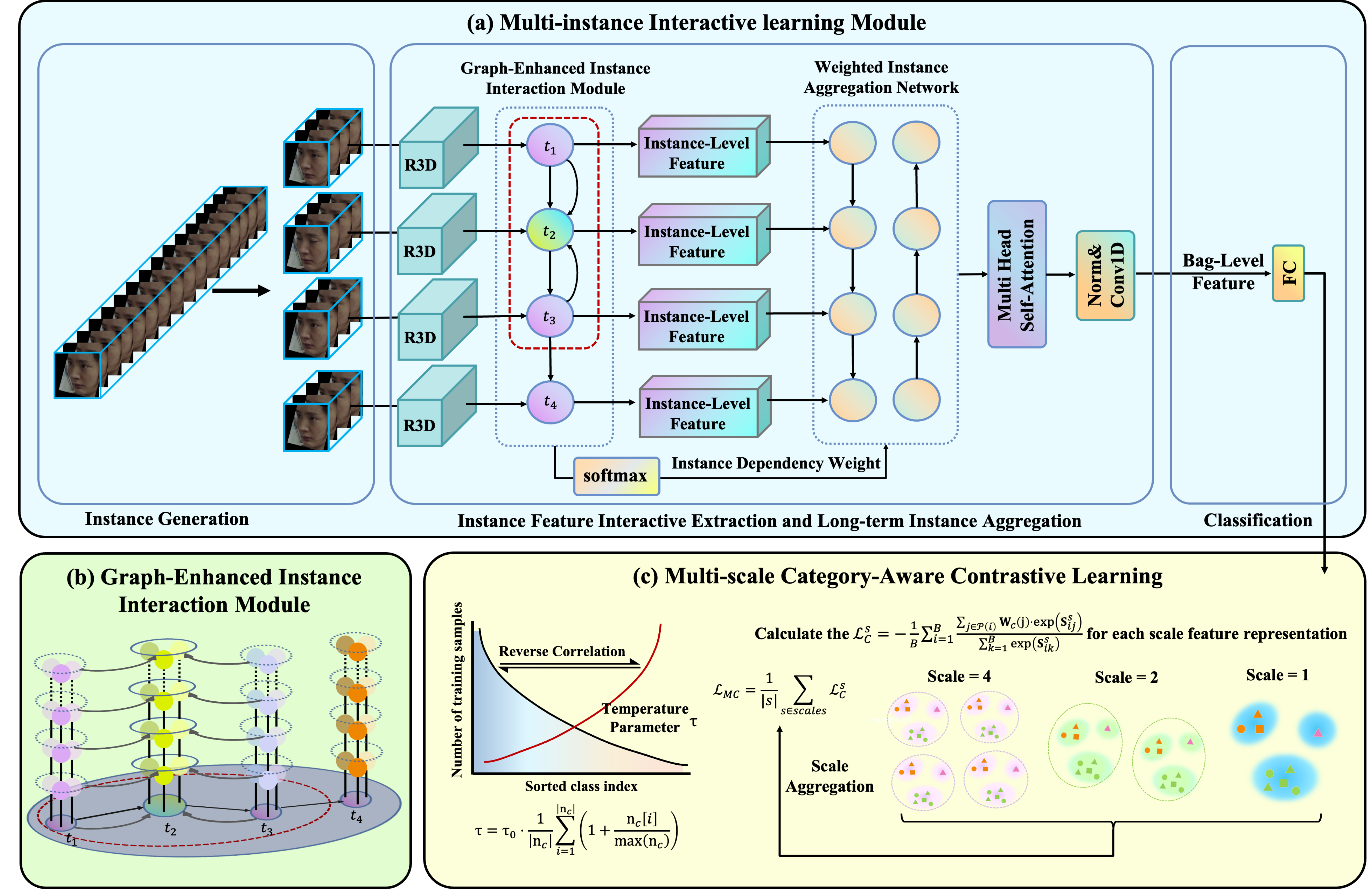} 
\caption{An overview of the proposed\name framework. (a) The flow of the proposed Multi-instance Interactive learning Module. (b) Schematic of the Graph-Enhanced Instance Interaction Module in\name. (c) Multi-scale Category-aware Contrastive Learning.}
\label{ours}
\end{figure*}

\section{RELATED WORK}

\subsection{Dynamic Facial Expression Recognition}

In recent years, with the development of deep learning technology, Dynamic Facial Expression Recognition (DFER) has made significant progress, demonstrating performance significantly superior to methods based on hand-crafted features \cite{Dhall_Goecke_Joshi_Sikka_Gedeon_2014,Liu_Shan_Wang_Chen_2014}. Some approaches first use CNNs \cite{He_Zhang_Ren_Sun_2016} to extract spatial features from each frame of a video, then employ RNNs (\eg, LSTM \cite{Hochreiter_Schmidhuber_1997} and GRU \cite{Chung_Gulcehre_Cho_Bengio_2014}) to capture temporal relationships between frames \cite{Zhang_Zheng_Cui_Zong_Li_2019,Ouyang_Kawaai_Goh_Shen_Ding_Ming_Huang_2017}. Additionally, methods based on 3D-CNNs have become a research hotspot \cite{Lee_Kim_Kim_Park_Sohn_2019}. By jointly learning spatial and temporal features, 3D-CNNs can simultaneously handle local features of each frame and global dynamic features. Studies such as CAER-Net \cite{Lee_Yoon} combine facial expressions with contextual information, while CP-Higher-Order Convolutions reduce computational complexity, making the application of 3D networks in DFER more efficient. 
Meanwhile, the introduction of transformer models offers new perspectives for DFER. For example, Former-DFER \cite{Zhao_Liu_2021} leverages convolutional spatial transformers and temporal transformers to separately learn spatial and temporal features; STT \cite{Ma_Sun_Li_2022} captures contextual relationships between frames through a transformer encoder; and NR-DFERNet \cite{Li_Sui_Zhu_zhao_2022} focuses on suppressing the interference of noisy frames in video sequences. Compared to traditional deep networks, these Transformer-based models demonstrate stronger global relationship modeling capabilities and greater potential for addressing diverse challenges. However, their large number of parameters and high computational cost limit their application in real-world scenarios. Moreover, while current DFER methods have made significant progress in capturing spatio-temporal features, handling noise, and addressing complex real-world conditions, they primarily focus on global feature modeling and optimizing inter-frame relationships, without specifically tailoring methods to the characteristics of expression categories. To address this, Li \etal \cite{Li_Niu_Zhu_Zhao_2023} proposed IAL and a GCA module, enhancing the classification of low-intensity expressions by adjusting the importance of feature map channels. However, this approach exacerbates the model’s neglect of long-tail categories \cite{10105457}. Long-tail categories have very few samples, making it challenging for the model to effectively learn the features of these categories and achieve balanced accuracy across all categories.

\subsection{Multi-Instance Learning}

Multi-Instance Learning (MIL) \cite{carbonneau2018multiple} is a technique designed to address problems with imprecise labels and has been widely applied in fields such as image processing and sentiment analysis. For example, Zhu \etal \cite{10.1007/978-3-031-72920-1_19} proposed DGR-MIL, which leverages cross-attention mechanisms to integrate global vectors, capturing the diversity of instances in MIL. Fu \etal \cite{9413640} introduced the MAEC model, which combines adversarial auto encoders with MIL to smooth category boundaries through regularization of latent representations and capture the most salient emotional moments through fine-grained speech segmentation. Li \etal \cite{10688167} developed a multi-instance causal learning model that integrates domain knowledge to construct causal graphs and identify latent causal relationships in multichannel physiological signals. Although these methods have achieved significant results in their respective fields, they have not been explored specifically for the DFER task. In the field of DFER, Wu \etal \cite{Chongliang} applied an MIL-based framework using a differentiable Hidden Markov Model (HMM) as a laboratory-controlled dynamic facial expression recognition classifier, with facial landmarks as input features. However, this method showed poor adaptability to “in-the-wild” DFER tasks. M3DFEL \cite{10204167} was the first to apply MIL to “in-the-wild” DFER and achieved promising results, but it failed to sufficiently address the inherent long-tailed category distribution in dynamic facial expressions, leading to significant inductive bias in the model.

\section{METHOD}
\subsection{Overview}

Dynamic Facial Expression Recognition (DFER) poses significant challenges due to intricate temporal dependencies and the long-tailed distribution of emotional categories. To address these challenges, we propose a novel framework, \name, which integrates multi-instance interaction and aggregation with category-aware multiscale contrastive learning. The \name framework is composed of three key components: the Graph-Enhanced Instance Interaction Module (GEIIM), the Weighted Instance Aggregation Network (WIAN), and the Multiscale Category-Aware Contrastive Learning (MCACL) loss. These components work in tandem to flexibly capture inter-frame dependencies and systematically address data imbalance, significantly enhancing both the robustness and generalization of the model.

\subsection{Multi-instance Interactive learning Module}
\textbf{Graph-Enhanced Instance Interaction Module.} 
To strengthen feature associations between neighboring instances in MIL, we introduce the Graph-Enhanced Instance Interaction Module (GEIIM). GEIIM uses a dynamic adjacency matrix to model complex contextual relationships among instances and employs graph convolution to facilitate efficient feature interaction. For input instance feature $\bf{X} \in \mathbb{R}^{B \times T \times C}$, $B$ represents the batch size, $T$ represents the number of instances, and $C$ represents the feature dimension. To construct the dependencies between frames, GEIIM employs learnable node embedding parameters $\mathbf{N}_1 \in \mathbb{R}^{T \times d}$ and $\mathbf{N}_2 \in \mathbb{R}^{d \times T}$, and generates the dynamic adjacency matrix $A$ through the following formula:
\begin{equation}
\textbf{A} = \text{Softmax}(\text{ReLU}(\mathbf{N}_1 \cdot \mathbf{N}_2)),
\end{equation}
the ReLU activation function suppresses irrelevant noise, and the Softmax operation normalizes the weights, which enhances the robustness of relation modeling by preventing excessive sensitivity to low-quality features, and finally ensures that the adjacency matrix effectively captures the dependencies between instances. Then, the constructed adjacency matrix $\textbf{A}$ is incorporated into a graph convolution operation to enable feature propagation among instances based on linear diffusion theory.  The feature interaction is updated as:
\begin{equation}
\bf{H} = \alpha \bf{X} + (1 - \alpha) \bf{A \cdot X},
\end{equation}
where $\alpha \in [0, 1]$ is a learnable coefficient that balances the fusion of original features $\bf{X}$ and the adjacency-enhanced features $\bf{X_A}$. This mechanism allows GEIIM to dynamically adjust the importance of individual instance information and neighborhood relationships, enhancing spatio-temporal interaction modeling.

\textbf{Weighted Instance Aggregation Network.} 
To enhance the quality of instance feature representation in MIL, we improve the traditional LSTM and propose the Weighted Instance Aggregation Network (WIAN). Compared to traditional LSTM, WIAN introduces a Dynamic Weight Gate (DWG) to evaluate the importance of each instance and achieve weighted feature fusion. The calculation of the DWG is defined as follows:  
\begin{equation}
\mathbf{d}_t = \sigma(\mathbf{W}_s \mathbf{x}_t + \mathbf{U}_s \mathbf{h}_{t-1}) \cdot \textbf{w}_t,
\end{equation}
where $\sigma$ is the Sigmoid function, and $\mathbf{s}_t$ is used to dynamically adjust the contribution of key instances, serving as the final representation of dynamic instance weights. Here, $\textbf{w}_t$ represents the instance weights extracted from the GEIIM, which are used to evaluate the importance of instances during feature aggregation, thereby highlighting key instances and suppressing noise interference. The calculation of $\textbf{w}_t$ is as follows:  
\begin{equation}
\textbf{w}_t = \text{Softmax}(\mathbf{H}_t, \text{dim} = -1),
\end{equation}
where $\mathbf{H}_t$ represents the feature matrix output from the GEIIM. The Softmax operation is applied along the last dimension to normalize the instance weights, making the weights of key instances more prominent while suppressing the weights of noisy instances. In WIAN, the dynamic weight $\mathbf{d}_t$ is used to dynamically adjust the contribution of the candidate cell state $\tilde{\mathbf{c}}_t$. During the feature aggregation stage, the updates for the cell state and hidden state are given by:  
\begin{equation}
\mathbf{c}_t = \mathbf{f}_t \cdot \mathbf{c}_{t-1} + \mathbf{i}_t \cdot \tilde{\mathbf{c}}_t \cdot \mathbf{d}_t, 
\end{equation}
\begin{equation}
\mathbf{h}_t = \mathbf{o}_t \cdot \tanh(\mathbf{c}_t),
\end{equation}
where $\mathbf{f}_t, \mathbf{i}_t, \mathbf{o}_t$ are the inherent forget, input, and output gates in LSTM. Through this mechanism, WIAN can precisely model and efficiently aggregate complex instance interactions in dynamic facial expression sequences. Furthermore, the features generated by WIAN are processed through a multi-head self-attention module, further modeling global temporal relationships. The WIAN module enables accurate modeling and efficient aggregation of complex inter-frame interactions, significantly enhancing the robustness and generalization capability of the model in dynamic facial expression recognition tasks.

\begin{table*}[t]
\renewcommand{\arraystretch}{1.3}
\centering
\setlength{\tabcolsep}{3mm}
\caption{Comparison (\%) of our\name with state-of-the-art methods on DFEW. * means that because MSCM has not released the source code, our table uses the experimental data under only fold 1 of DFEW in its paper. (\textbf{Bold}: Best, \underline{Underline}: Second best)}
\scalebox{1.0}{
\begin{tabular}{ c|ccccccc|cc|c }
\hline
\multirow{2}{*}{\textbf{Method}} & \multicolumn{7}{c|}{\textbf{Accuracy of Each Emotion(\%)}} & \multicolumn{2}{c|}{\textbf{Metrics (\%)}} & \multirow{2}{*}{\makecell[c]{\textbf{FLOPs}\\\textbf{(G)}}} \\ 
\cmidrule(lr){2-10}
& \textbf{Hap.}	  & \textbf{Sad.}    & \textbf{Neu.}   & \textbf{Ang.}	   & \textbf{Sur.}	& \textbf{Dis.}	  & \textbf{Fea.}	 & \textbf{WAR}   	& \textbf{UAR} \\ \hline
ResNet18+LSTM \cite{dfew} &78.00 &40.65 &53.77 &56.83 &45.00 &\underline{4.14} &21.62 &53.08 &42.86 &7.78\\
EC-STFL \cite{dfew}  &79.18 &49.05 &57.85 &60.98 &46.15 &2.76 &21.51 &56.51 &45.35 &8.32\\
Former-DFER \cite{Zhao_Liu_2021}   &84.05	 &62.57	   &67.52	&70.03	  &56.43	&3.45	 &31.78	    &65.70	   &53.69	 &9.11 \\ 
STT \cite{Ma_Sun_Li_2022}          &87.36	 &67.90   &64.97	&71.27	  &53.10	&3.49	 &\underline{34.04}	    &66.45	   &54.58	 & N/A \\ 
NR-DFERNet \cite{Li_Sui_Zhu_zhao_2022}    &88.47	 &64.84	   &\underline{70.03}	& \underline{75.09} &\underline{61.60}	&0.00	 &19.43	    &68.19	   &54.21	 &6.33 \\
LOGO-Former \cite{Ma2023LogoFormerLS}          &85.39	 & 66.52	   &68.94	& 71.33 	  &54.59	&0.00	 &32.71	    &66.98 &54.21 &N/A \\ 
GCA+IAL \cite{Li_Niu_Zhu_Zhao_2023}           &87.95	 &67.21	   &\textbf{70.10}	&\textbf{76.06}	  &\textbf{62.22}	&0.00	 &26.44	    &69.24	   &55.71	 &9.63  \\ 
M3DFEL \cite{10204167}    &\textbf{89.59} 	 &\underline{68.38}  &67.88	&74.24	  &59.69	&0.00	 &31.63  	& \underline{69.25}	   &56.10	 &\textbf{1.65}  \\ 
T-MEP \cite{10250883}   & N/A	 & N/A  & N/A  & N/A  & N/A  & N/A  & N/A  	&68.85	   &\underline{57.16}	 &N/A  \\ \cmidrule(lr){1-11}
\name (Ours)    &\underline{89.22} &\textbf{69.48} & 63.72 & 73.47 & 60.92  & \textbf{29.65} &\textbf{42.57} &\textbf{69.56}  &\textbf{63.21} &\underline{1.69}\\ \hline
\end{tabular}
}

\label{res-dfew}

\end{table*}

\begin{table}[h]
\renewcommand{\arraystretch}{1.3}
\centering
\setlength{\tabcolsep}{3mm}
\caption{Comparison (\%) of our\name with the state-of-the-art methods on FERV39K.}
\scalebox{1.0}{
\begin{tabular}{ c|cc }
\hline
 
 \textbf{Method} & \textbf{WAR}   & \textbf{UAR} \\ \hline
 
2ResNet18+LSTM \cite{ferv39k} &43.20 &31.28 \\
NR-DFERNet \cite{Li_Sui_Zhu_zhao_2022}  &45.97	&33.99 \\
Former-DFER \cite{Zhao_Liu_2021} &46.85	&37.20	 \\
3D-DSwin \cite{10394270}    &46.98 &37.66  \\
M3DFEL \cite{10204167} &47.67	&35.94	 \\ 
LOGO-Former \cite{Ma2023LogoFormerLS}     &\underline{48.13}	&\underline{38.22} \\
 
\cmidrule(lr){1-3} 
\name (Ours)	& \textbf{48.57}	& \textbf{40.25}	 \\ \hline
\end{tabular}
}

\label{res-39k}

\end{table}

\begin{table}[t]
\setlength{\tabcolsep}{1.0mm}
\renewcommand{\arraystretch}{1.1}
\centering
\caption{The importance of GEIIM, WIAN, MCCL in the proposed\name framework.}
\scalebox{1.0}{
\begin{tabular}{c|ccc|cc}
\hline
\multirow{2}{*}{Setting} & \multicolumn{3}{c|}{Method} & \multicolumn{2}{c}{DFEW (\%)}   \\ \cmidrule(lr){2-6} 
  & GEIIM  & WIAN     & MCCL   & WAR   & UAR     \\ \hline
a & \ding{55} & \ding{55}   & \ding{55}  & 64.55 & 54.17 \\  \cmidrule(lr){1-6} 
b  & \CheckmarkBold & \ding{55}   & \ding{55}  & 68.11 & 57.39 \\ 
c  & \CheckmarkBold   & \CheckmarkBold   & \ding{55}   &68.62 & 56.26\\ 
d  & \CheckmarkBold  & \CheckmarkBold   & \CheckmarkBold &\textbf{69.91} & \textbf{64.34} \\  \hline
\end{tabular}
}

\label{ablation}
\end{table}

\subsection{Multiscale Category-Aware Contrastive Learning}

To effectively address the challenges of category imbalance and complex feature distributions in long-tailed dynamic facial expression recognition, we propose the Multiscale Category-aware Contrastive Learning (MCCL) loss. MCCL enhances model performance in long-tailed distributions by incorporating dynamic category-aware weighting and multiscale feature aggregation mechanisms, improving both tail-category recognition and overall robustness. For an input feature $\mathbf{X}' \in \mathbb{R}^{B \times C}$, MCCL combines category imbalance and sample difficulty information to generate dynamic category-aware weights $\mathbf{W}_\text{c}$:
\begin{equation}
\mathbf{W}_\text{c}(i) = \frac{1}{\text{n}_c[i]} \cdot \big(1 - \text{softmax}(\mathbf{y}_\text{p})[i]\big),
\end{equation}
where $\text{n}_c[i]$ denotes the number of samples in category $i$, and $\mathbf{y}_\text{p}$ represents the model’s predicted probability distribution. By dynamically adjusting the importance of tail categories and hard-to-classify samples, $\mathbf{w}_\text{c}$ provides stronger learning signals. To capture features across multiple scales, MCCL performs hierarchical representation of input features:
\begin{equation}
\mathbf{X}'_s = \text{Proj}_s(\text{Pool}_s(\mathbf{X'})).
\end{equation}
where $\text{Pool}_s$ and $\text{Proj}_s$ represent adaptive pooling and linear projection operations on scale $s$, respectively. For each scale $s$, the corresponding feature representation $\mathbf{X}'_s$ is used to compute the similarity matrix:
\begin{equation}
\mathbf{S}_{ij}^s = \frac{\mathbf{X}'_s[i] \cdot \mathbf{X}'_s[j]}{\tau \|\mathbf{X}'_s[i]\| \|\mathbf{X}'_s[j]\|},
\end{equation}
with the temperature parameter $\tau$ dynamically adjusted as:
\begin{equation}
\tau = \tau_0 \cdot \frac{1}{|\text{n}_c|} \sum_{i=1}^{|\text{n}_c|} \Big(1 + \frac{\text{n}_c[i]}{\max(\text{n}_c)}\Big),
\end{equation}
to reduce the influence of frequent categories on similarity computation and emphasize tail-category features. The contrastive loss for scale $s$ is defined as:
\begin{equation}
\mathcal{L}_\text{c}^s = - \frac{1}{B} \sum_{i=1}^{B} \frac{\sum_{j \in \mathcal{P}(i)} \mathbf{W}_\text{c}(i) \cdot \exp(\mathbf{S}_{ij}^s)}{\sum_{k=1}^{B} \exp(\mathbf{S}_{ik}^s)},
\end{equation}
where $\mathcal{P}(i)$ denotes the set of positive samples of the same category as sample $i$. The overall contrastive loss is the average across all scales:
\begin{equation}
\mathcal{L}_\text{MC} = \frac{1}{|S|} \sum_{s \in \text{scales}} \mathcal{L}_\text{c}^s.
\end{equation}

To further improve feature robustness and generalization in supervised classification tasks, MCCL introduces the Classification Enhancement Term (CET), which integrates feature augmentation regularization and intensity-aware loss \cite{Li_Niu_Zhu_Zhao_2023} for maximum confusion classification. the total MCCL loss is expressed as:
\begin{equation}
\mathcal{L}_\text{all} = \mathcal{L}_\text{MC} + \mathcal{L}_\text{CET}.
\end{equation}
MCCL achieves optimized contrastive learning under long-tailed distributions, significantly improving tail-category recognition performance, and enhancing the overall generalization of the model. The details of $\mathcal{L}_\text{CET}$ are provided in the Appendix.

\section{EXPERIMENTS}
\subsection{Implementation Details and Evaluation Metrics}
Our experiment was performed on an RTX A6000 GPUs, and all face images are resized to 112×112 in the experiment. For each video, 16 frames are extracted as samples. The feature extraction network uses the standard R3D model with pre-trained weights from Torchvision. The model is trained using the AdamW optimizer and cosine scheduler for 300 epochs. The learning rate is set to 4e-4, with a minimum learning rate of 3e-6, and the weight decay is set to 0.05.

Training datasets include DFEW \cite{dfew} and FERV39k \cite{ferv39k}. DFEW contains 16,000+ video clips from over 1,500 films, annotated by 10 professionals into 7 basic expressions. FERV39K, the largest DFER dataset, includes 38,935 clips across 22 sub-scenes, annotated by 30 professionals. 

We use weighted average recall (WAR) and unweighted average recall (UAR) as metrics. WAR measures the model's overall classification accuracy, with a stronger influence from majority classes, making it suitable for assessing real-world applicability. In contrast, UAR is unaffected by class distribution and better reflects the model's balanced performance across categories.
 
\subsection{Comparison with the State-of-the-art Methods}
Consistent with previous work, experiments are conducted under 5-fold cross-validation on DFEW. The experimental results are shown on Tab.~\ref{res-dfew}. It can be seen that our method achieves the best results both in WAR and in UAR. As shown in Tab.~\ref{res-39k}, our method achieves good results in both WAR and UAR on FERV39k. Our method significantly outperforms previous methods in categories with smaller sample sizes. This further validates the effectiveness of our approach for overall performance and small sample size categories.


\subsection{Ablation Studies and Visualization}
\begin{figure}[t]
\centering
\includegraphics[width=0.48\textwidth]{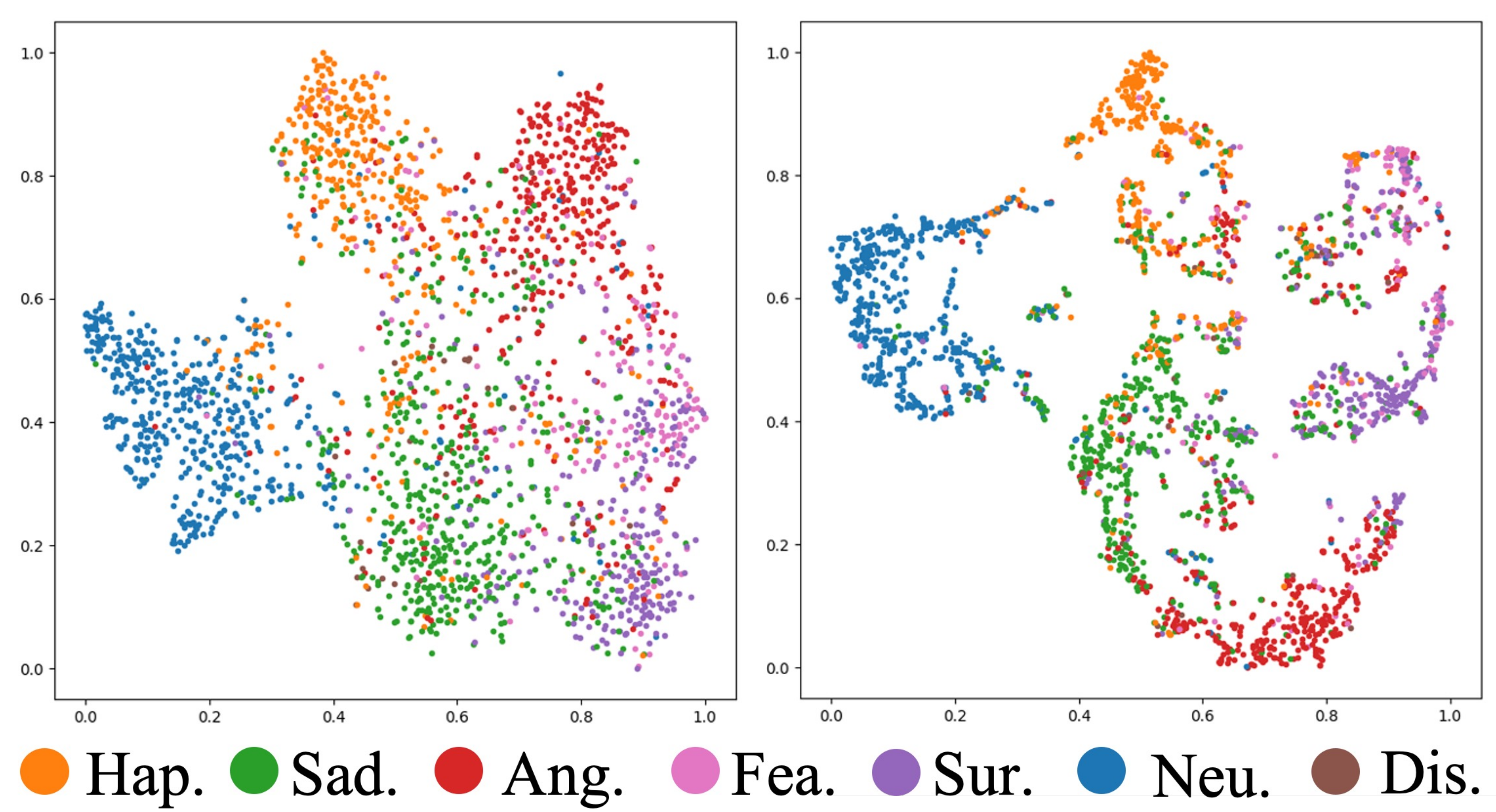} 
\caption{Visualization of the feature distribution learned by the baseline (left) and our method (right) on DFEW.}
\label{sne}
\end{figure}

We conduct ablation studies on DFEW Flod 5 to demonstrate the effectiveness of each component (\ie, GEIIM, WIAN, MCCL) of our method. In our experiments, the cross-entropy loss is a baseline loss. R3D, Self-Attention, and Conv1D are combined as baseline model. Meanwhile, we use t-SNE \cite{SNE:v9:vandermaaten08a} to analyze the feature distributions learned by the baseline and our method, as shown in Fig.~\ref{sne}. This effectively proves that our method can effectively interact with instances and aggregate time-series information, and achieve balanced feature contrastive learning, making feature distributions of the same categories tighter and the boundaries between different categories more distinct compared to the baseline method. This demonstrates that\name effectively guides the model to focus on and aggregate individual categories, performing well in categories with small sample sizes.

\section{Conclusion}
We propose a novel framework,\name, to enhance DFER performance by addressing the challenges posed by long-tailed category distributions and the complexity of spatio-temporal feature modeling. By integrating spatio-temporal dependency modeling ability of MIL with long-tailed optimization, our framework overcomes the limitations of existing methods, which often struggle with poor generalization and biased recognition of tail expression categories. Specifically, we introduce the Graph-Enhanced Instance Interaction Module (GEIIM), which captures intricate spatio-temporal relationships through adaptive adjacency matrices and multiscale convolutions, thereby improving the modeling of spatio-temporal features. To enhance instance-level feature aggregation, we develop the Weighted Instance Aggregation Network (WIAN), which dynamically assigns weights based on instance importance. Furthermore, we introduce a Multiscale Category-aware Contrastive Learning (MCCL) loss to balance training between head and tail categories, effectively mitigating the impact of category imbalance. In future work, we plan to further optimize spatio-temporal modeling techniques and explore more advanced strategies for handling long-tailed categories, ultimately constructing a more efficient and robust DFER solution.

\section*{ACKNOWLEDGEMENTS}
This work is supported by Students' Innovation and Entrepreneurship Foundation of University of Science and Technology of China (Grant No. CY2024X019A).

\bibliographystyle{IEEEbib}
\bibliography{icme2025references}

\end{document}